\documentclass[letterpaper, 10 pt, conference]{ieeeconf}  

\IEEEoverridecommandlockouts                              

\usepackage{xspace}
\usepackage{booktabs}
\usepackage{multirow}
\usepackage{graphicx}
\usepackage{amsmath}
\usepackage{array}
\usepackage{ragged2e}
\usepackage{amsfonts}
\usepackage{adjustbox}
\usepackage{pifont}
\usepackage{algorithm}
\usepackage{algpseudocode}
\usepackage{array}

\newcolumntype{L}[1]{>{\raggedright\arraybackslash}p{#1}}
\newcolumntype{C}[1]{>{\centering\arraybackslash}p{#1}}
\newcommand{\oursol}{\textsc{RoboBRIDGE}\xspace}

\title{\LARGE \bf
\oursol: A Modular Framework for Bridging Policies to Robust Real-World Robotic Agents
}

\author{Sihyung Yoon$^{1,*}$, Minjong Yoo$^{1, *}$, Sanghyun Ahn$^{1}$, Seojeong Choi$^{1}$, Honguk Woo$^{1, \dagger}$ 
\thanks{
This work was supported by 
Institute of Information \& communications Technology Planning \& Evaluation (IITP) grant funded by the Korea government (MSIT), (RS-2025-25442569, AI Star Fellowship Support Program (Sungkyunkwan Univ.), RS-2026-25528384, Resource-Intensive AI Technologies Based on Sustainable GPU Integrated Platforms), 
the National Research Foundation of Korea (NRF) grant funded by the Korea government (MSIT) (No. RS-2026-25474409), 
IITP-ITRC (Information Technology Research Center) grant funded by the Korea government (MSIT) (IITP-2025-RS-2024-00437633).
}
\thanks{$^{1}$Department of Computer Science and Engineering, Sungkyunkwan University}%
\thanks{$^{*}$Equal contribution.}
\thanks{$^{\dagger}$Honguk Woo is the corresponding author.}
}
\begin{document}

\maketitle
\thispagestyle{empty}
\pagestyle{empty}

\begin{abstract}
Vision-Language-Action (VLA) models have attracted growing interest as a scalable approach to robotic manipulation. While these models are effective action predictors, deploying them as robotic agents exposes critical gaps: no mechanism for failure recovery, inconsistent execution over long horizons, and limited robustness to shifts in observations, tasks, or embodiments. 
Existing solutions address these limitations individually through model retraining or environment-specific modules, yet what is needed is a general framework that systematically transforms a pretrained VLA into a robotic agent. 
We present \oursol, a modular framework that provides an orchestration layer over five coordinated modules, namely Monitor, Perceptor, Planner, Controller, and Robot Interface, to compose robust robotic agents from off-the-shelf components, including pretrained VLAs. The Monitor pairs rapid failure detection with hierarchical recovery to correct errors before they cascade.
When the environment diverges from the current plan, the Planner triggers replanning while the Perceptor updates scene understanding asynchronously, avoiding execution stalls. Within the Controller, primitive skill fine-tuning factors manipulation into domain-invariant primitives with dedicated LoRA adapters, reducing sensitivity to domain shifts when a VLA is used. 
Across LIBERO, RoboCasa, and real-world case studies spanning multiple robot platforms and VLA backbones, \oursol consistently outperforms both standalone policies and prior augmented VLA deployments. These results suggest that reliable robotic agency does not arise from scaling action predictors alone, but from structured orchestration around them.
\end{abstract}

\section{INTRODUCTION}

Vision-Language-Action (VLA) models have emerged as promising foundation models for robotic manipulation, mapping multimodal observations and language instructions directly to low-level actions with increasingly broad task coverage~\cite{kim2024openvla, black2025pi05, nvidia2025grootn15, shukor2025smolvla}.
Yet a capable action predictor does not, by itself, constitute a robotic agent, which must further detect and recover from execution failures, maintain consistent behavior over long task horizons, and generalize across shifts in observations, tasks, and embodiments~\cite{zhong2025survey, sapkota2025vision}.
Current VLA models, deployed as monolithic forward-pass policies, provide none of these capabilities: a missed grasp goes unnoticed, compounding errors destabilize multi-step plans, and performance degrades sharply when the deployment domain departs from the training distribution. 

\begin{figure}[t]
    \centering
    \includegraphics[width=0.90\columnwidth]{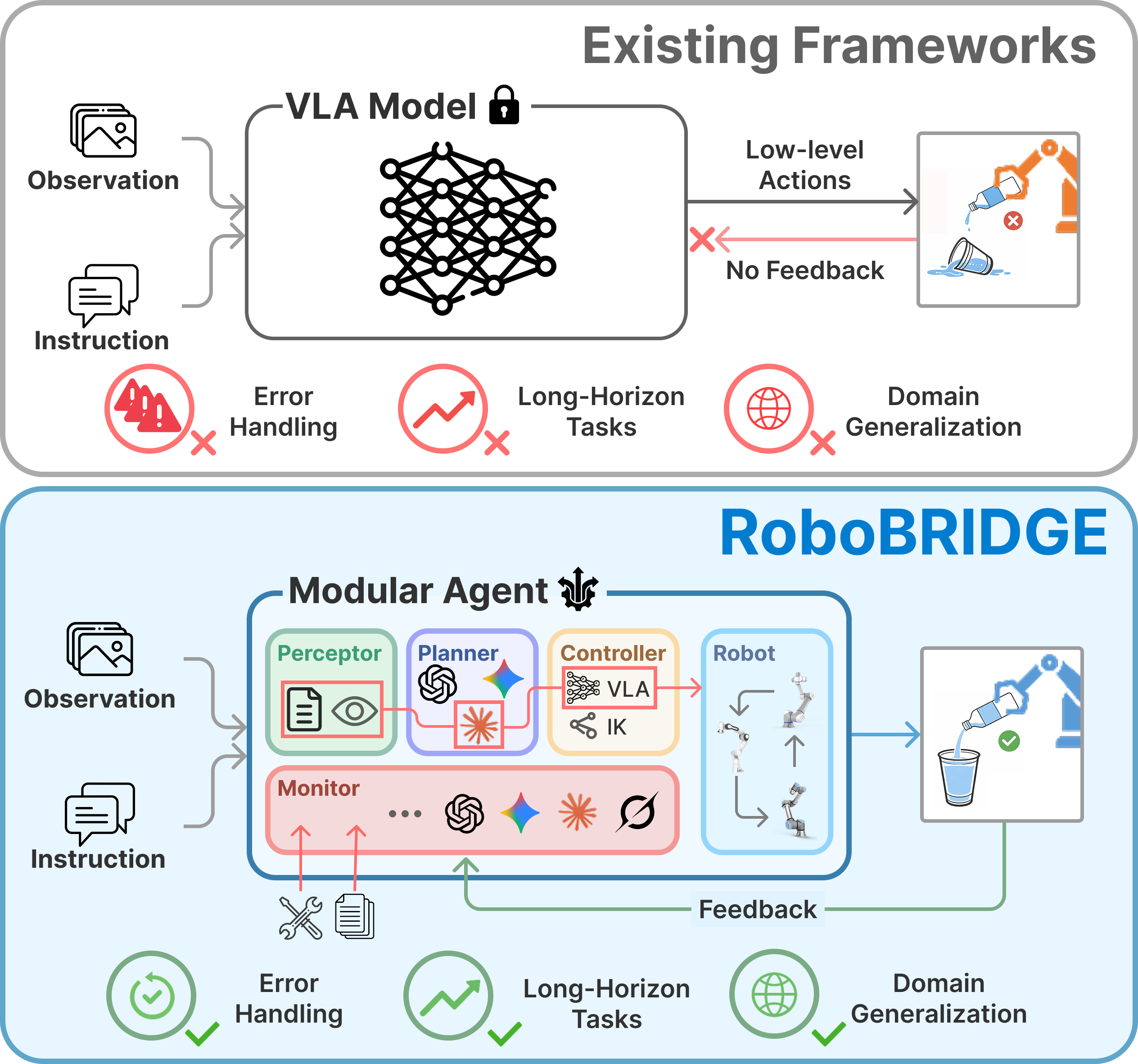}
    \caption{Existing VLA deployments execute actions in an open-loop manner and lack mechanisms for error handling, long horizon consistency, and domain generalization. \oursol wraps an action generating policy with modular perception, planning, monitoring, and robot interface components to close the loop with feedback and enable robust execution.}
    \label{fig:Concept of RoboBRIDGE}
    \vspace{-0.15in}
\end{figure}

Several lines of work have addressed individual aspects of these limitations, including runtime monitoring for failure detection~\cite{zhou2025code, safe2025}, plan-execution misalignment recovery through grounded replanning~\cite{guo2024doremi, cyclevla2026}, and LLM-guided task decomposition with skill grounding~\cite{ahn2022can, zhou2024llmbt, zhou2024isrllm}, and cross-domain policy training with parameter-efficient adaptation for domain robustness~\cite{kim2024openvla, kachaev2025don}.
However, these efforts remain fragmented: each targets a specific failure mode and assumes a particular environment setting, leaving no unified solution.
We argue that such a framework must be policy-agnostic, capable of wrapping an arbitrary action-generating controller without retraining or architectural changes.

\begin{figure*}[t]
    \centering
    \begin{adjustbox}{width=.95\textwidth}
    \includegraphics[width=\textwidth]{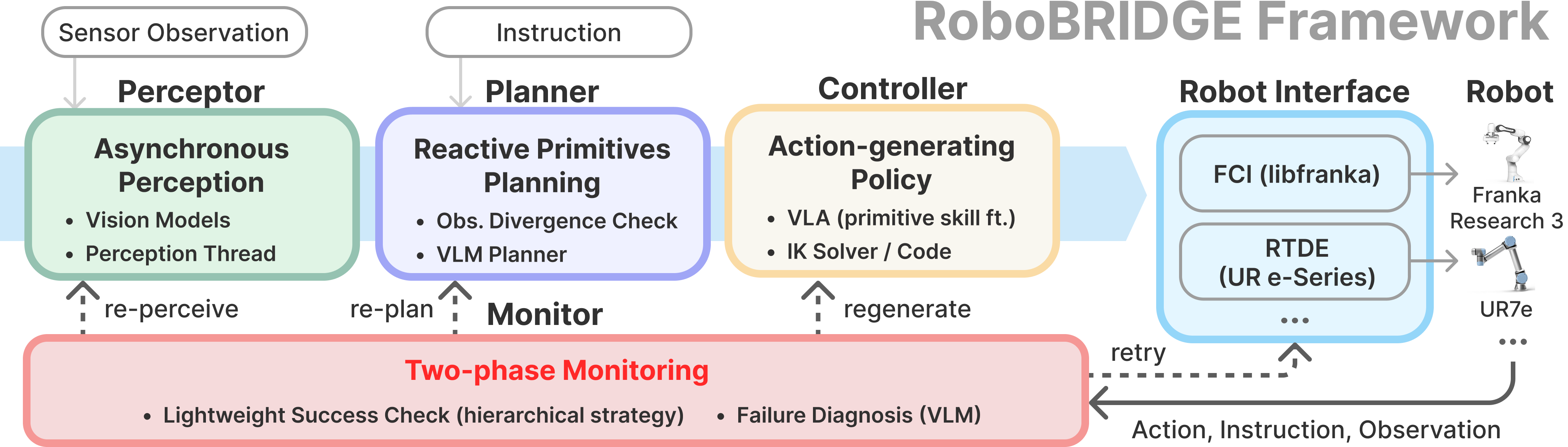}
    \end{adjustbox}
    \vspace{-0.05in}
    \caption{Overall framework of \oursol. The Perceptor updates an object centric scene state asynchronously, and the Planner generates reactive primitive skills with divergence triggered replanning. The Controller executes primitives with an action generating policy, while the Robot Interface abstracts robot specific APIs. A two-phase Monitor performs success checking and failure diagnosis, invoking hierarchical recovery such as retry, regenerate, replan, and re-perceive.}
    \label{fig:overall_framework}
    \vspace{-0.15in}
\end{figure*}

We draw a parallel to how Large Language Models (LLMs) evolved from isolated text generators into versatile agents.
Raw LLMs, despite their strong generative capabilities, could not verify their own outputs, reliably perform multi-step tasks, or generalize across unseen domains.
The community addressed this not by training ever-larger models, but by building general-purpose orchestration frameworks~\cite{ langchain_website, wu2024autogen} that wrap any LLM with tool use, planning, and verification modules, transforming a base model into an agent without modifying its weights.
We argue that VLA-based robotic manipulation faces an analogous bottleneck: the missing piece is not a better policy, but a general framework that elevates any policy into a robotic agent.

We present \oursol, a modular, generalizable orchestration framework that converts any action-generating policy into a robust robotic agent.
As illustrated in Figure~\ref{fig:overall_framework}, the framework is composed of five coordinated modules: \textbf{Monitor}, \textbf{Perceptor}, \textbf{Planner}, \textbf{Controller}, and \textbf{Robot Interface}.
Any policy, including but not limited to VLAs, occupies the Controller slot, while the surrounding modules supply the capabilities that monolithic deployment lacks.
Each of the three identified limitations is addressed by a dedicated mechanism.
First, \textbf{two-phase monitoring} continuously evaluates execution outcomes and, upon detecting failures, selects among hierarchical recovery strategies, enabling prompt correction before errors cascade.
Second, \textbf{reactive planning with asynchronous perception} decouples scene understanding from execution into concurrent threads, triggering replanning only when the environment diverges materially from the current plan, thereby maintaining consistent task progression over long horizons.
When a VLA serves as the controller, we further introduce \textbf{primitive skill fine-tuning}, which factors manipulation into domain-invariant primitives (e.g., move, grip, ...) with dedicated LoRA adapters, reducing sensitivity to domain shifts and enabling robust sim-to-real and cross-robot transfer.

We validate \oursol in both simulation and real-world settings using three VLA backbones: SmolVLA~\cite{shukor2025smolvla}, $\pi_{0.5}$~\cite{black2025pi05}, and GR00T-N1.5~\cite{nvidia2025grootn15}.
In simulation, we evaluate on LIBERO and RoboCasa~\cite{robocasa2024} across diverse manipulation tasks, where \oursol consistently improves standalone and augmented VLA baselines. In real-world trials across multiple robot platforms and tasks, \oursol yields robust gains under embodiment and environmental variability, enabled by failure-aware monitoring and recovery.

Our contributions are summarized as follows:
\begin{itemize}
\item We propose \oursol, a modular, generalizable orchestration framework that transforms any action-generating policy into a robotic agent by providing failure recovery, long-horizon consistency, and cross-domain robustness through five coordinated modules.
\item When a VLA serves as the controller, we further introduce primitive skill fine-tuning, which factors manipulation into domain-invariant primitives with dedicated LoRA adapters, reducing performance degradation under domain shifts.
\item We validate \oursol on RoboCasa and real-world episodes across several VLA backbones, demonstrating that the framework yields consistent improvements regardless of the underlying policy.
\end{itemize}

\section{Related Work}
\textbf{VLA models.}
The trajectory from task-specific controllers to generalist policies has been driven by large-scale data and model scaling.
RT-1~\cite{brohan2022rt} demonstrates that a single Transformer can absorb hundreds of real-world tasks, DROID~\cite{droid2024} broadens distributional coverage beyond curated lab settings, and the Open X-Embodiment effort~\cite{openx2024} aggregates heterogeneous robot datasets under a unified training recipe.
VLA models extend this trend by unifying vision, language, and action prediction into a single architecture, with Octo~\cite{octo2024} demonstrating cross-embodiment reuse, $\pi_{0.5}$~\cite{black2025pi05} scaling to dexterous manipulation, GR00T~\cite{nvidia2025grootn15} targeting humanoid-scale control, and SmolVLA~\cite{shukor2025smolvla} exploring compact deployment.
Despite these advances, VLA models remain action predictors rather than robotic agents: they lack mechanisms to detect execution failures, maintain consistency over long horizons, or adapt to domain shifts at inference time.
\oursol treats any such model as a pluggable controller module, supplying the surrounding infrastructure needed to elevate it into a robotic agent.

\textbf{Augmented VLA deployment.}
A growing body of work has improved the reliability of learned policies by addressing specific failure modes in deployment.
For failure detection and recovery, Code-as-Monitor~\cite{zhou2025code} programs constraint-aware runtime checks, DoReMi~\cite{guo2024doremi} detects plan-execution misalignment and triggers grounded replanning, SAFE~\cite{safe2025} trains a multitask failure detector for VLA outputs, FailSafe~\cite{failsafe2025} learns recovery behaviors from failure-annotated demonstrations, and CycleVLA~\cite{cyclevla2026} augments a VLA with subtask backtracking and self-correction decoding.
For long-horizon task decomposition, SayCan~\cite{ahn2022can} couples LLM reasoning with pretrained skills and affordance scoring, LLM-BT~\cite{zhou2024llmbt} translates instructions into dynamically updated behavior trees, and ISR-LLM~\cite{zhou2024isrllm} iteratively refines LLM-generated plans with explicit validation.
For domain robustness, prior work has explored domain randomization~\cite{kim2024openvla, openx2024}, sim-to-real transfer pipelines, and embodiment-specific fine-tuning strategies~\cite{kachaev2025don}.
However, these approaches each target a single failure mode and are tied to particular model architectures or environmental settings.
Rather than engineering a dedicated solution for each failure mode, \oursol provides a general orchestration framework in which standardized modules for perception, planning, monitoring, and control can be freely composed, enabling practitioners to equip any policy with the capabilities it lacks.

\textbf{Tool-augmented agents.}
The LLM community has demonstrated that augmenting a base model with external tools, memory, and verification yields more robust agents than scaling the model alone.
LATM~\cite{cai2024latm} synthesizes reusable tools for deterministic sub-task execution, AvaTaR~\cite{wu2024avatar} optimizes tool-use decisions through contrastive trajectory reasoning, and general-purpose orchestration frameworks~\cite{langchain_website, wu2024autogen} standardize how any LLM is wrapped with planning, tool invocation, and self-correction modules to form an agent.
In embodied settings, VoxPoser~\cite{huang2023voxposer} combines LLM-inferred constraints with VLM-generated 3D value maps, and ALRM~\cite{santos2026alrm} integrates code generation with tool-based execution for robotic control.
Yet no existing system provides a general orchestration framework for robotic manipulation that is agnostic to the underlying controller while jointly supplying monitoring, recovery, and replanning.
\oursol instantiates this role for robotic manipulation, transforming any action-generating policy into a robust robotic agent through a modular, coordinated control stack.

\section{APPROACH}

VLA models are effective action predictors but lack the essential properties of robust robotic agents.
Deployed as monolithic forward-pass policies, they provide no mechanism to monitor execution outcomes or initiate recovery when actions fail, causing a single missed grasp or collision to cascade through subsequent steps.
Each inference step is independent, offering no built-in means to track task progress or correct compounding errors over long horizons.
Performance further degrades under distribution shifts in observations, tasks, or embodiments, leaving sim-to-real transfer and cross-robot deployment inherently brittle.
\oursol addresses these limitations by wrapping a pretrained VLA inside a modular, tool-augmented control stack that preserves the model's learned manipulation capability while supplying failure recovery, long-horizon consistency, and domain robustness at inference time, without retraining the base model.

The framework comprises five modules, illustrated in Figure~\ref{fig:overall_framework}.
The \textbf{Perceptor} converts raw sensory observations into an object-centric state comprising identities, 3D poses, and semantic attributes such as affordances and task-relevant properties.
The \textbf{Planner} receives this refined state together with a language instruction and decomposes the task into an ordered sequence of primitive skills parameterized by additional context.
The \textbf{Controller} is the slot occupied by the action-generating policy, translating each primitive into low-level actions.
The \textbf{Robot Interface} abstracts embodiment-specific APIs, timing, safety limits, and coordinate transforms, allowing the same upstream logic to run across robots and simulators.
The \textbf{Monitor} closes the loop by verifying execution outcomes and triggering hierarchical recovery when failures are detected.

These modules jointly enable two core mechanisms.
(a)~\textbf{Two-phase monitoring} (Sec.~\ref{sec:monitor}) pairs a lightweight success check with detailed failure diagnosis, enabling prompt, targeted recovery without restarting from scratch.
(b)~\textbf{Reactive planning with asynchronous perception} (Sec.~\ref{sec:async}) decouples perception and execution into concurrent threads, triggering replanning only when the environment diverges materially from the current plan.
When a VLA serves as the controller, we further introduce (c)~\textbf{Primitive skill fine-tuning} (Sec.~\ref{sec:gpr}), which factors manipulation into domain-invariant primitives with dedicated LoRA adapters on the VLA backbone, reducing sensitivity to observation, task, and embodiment shifts.

\subsection{Two-phase Monitoring for Failure Recovery} \label{sec:monitor}

\begin{figure}[t]
    \centering
    \begin{adjustbox}{width=.95\columnwidth}
    \includegraphics[width=\columnwidth]{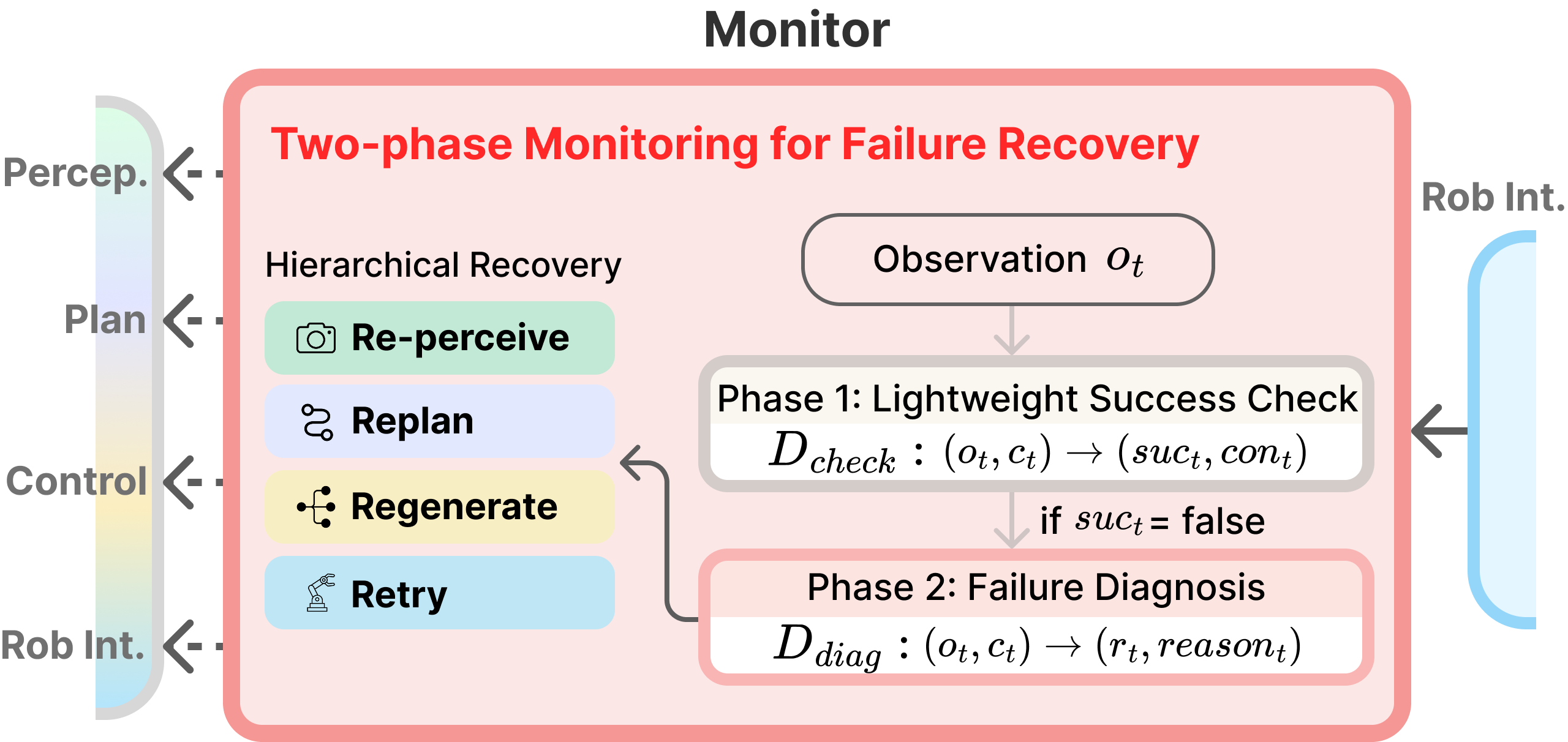}
    \end{adjustbox}
    \caption{Two-phase monitoring for failure recovery. A lightweight success check flags failure, then a diagnosis module selects a hierarchical recovery action from retry to re-perceive.}
    \label{fig:Two-phase Monitoring for Failure Recovery}
\end{figure}

Execution failures such as missed grasps, collisions, or kinematic infeasibility must be detected promptly to prevent cascading errors in long-horizon manipulation. Since most control steps succeed, running a full diagnostic at every cycle would introduce unnecessary latency. As illustrated in Figure~\ref{fig:Two-phase Monitoring for Failure Recovery}, we therefore separate fast success checking from detailed failure diagnosis, invoking the latter only when a failure is detected.

\textbf{Phase 1: Lightweight success check.} During execution, a success check model $D_{\text{check}}$, such as a vision-language model, periodically evaluates the current observation $o_t$ against the expected plan context $c_t$, producing a binary success flag $suc_t$ and a confidence score $con_t$: 
\begin{equation} 
D_{\text{check}}: (o_t, c_t) \mapsto (suc_t, con_t) 
\end{equation} 
To minimize latency, the query uses a constrained output format that suppresses extended reasoning, keeping the check lightweight enough to operate continuously alongside execution without introducing blocking overhead.

\textbf{Phase 2: Failure diagnosis and recovery.} When a failure is detected with high confidence ($suc_t{=}\text{false}$ and $con_t \ge \gamma_{\text{thresh}}$), the system immediately halts the robot and invokes a diagnosis model $D_{\text{diag}}$ that infers the root cause and selects a recovery target $r_t$ and recovery reason $reason_t$: 
\begin{equation} 
D_{\text{diag}}: (o_t, c_t) \mapsto (r_t, reason_t). 
\end{equation} 
The recovery target $r_t$ maps to a hierarchical strategy with four levels, ordered by increasing scope of re-computation: (1)~\textit{retry}, which re-executes the current primitive; (2)~\textit{regenerate}, which regenerates the trajectory while retaining the current plan and perception state; (3)~\textit{replan}, which replans from the latest asynchronous perception (Sec.~\ref{sec:async}); and (4)~\textit{re-perceive}, which forces a full re-perception before replanning when the object state is deemed unreliable. This hierarchy helps the system apply the least costly correction sufficient to resolve the failure, avoiding unnecessary re-computation while still enabling deep recovery when needed.

\subsection{Reactive Planning with Asynchronous Perception}
\label{sec:async}

Conventional manipulation pipelines run perception once before planning, implicitly assuming a static scene throughout execution. In practice, the robot's own actions and external disturbances frequently invalidate this assumption. We address this by decoupling perception from planning via an asynchronous producer-consumer architecture that hides perception latency behind execution and triggers replanning only when the scene changes materially.

\textbf{Asynchronous perception.}
Our system runs two concurrent threads connected through a thread-safe buffer $\mathcal{B}$. The \textit{perception thread} continuously updates $\mathcal{B}$ with the latest detection result $\bar{o}_t = D_{\text{percept}}(o_t)$ at a rate determined by inference speed. Since only the most recent observation is needed, $\mathcal{B}$ stores a single result rather than a full history.

\textbf{Divergence-triggered replanning.}
The \textit{execution thread} runs primitives sequentially. After each primitive completes, it compares the perception observation $\bar{o}_{\text{plan}}$ used for the current plan with the latest observation $\bar{o}_{lat}$ from $\mathcal{B}$. The divergence measure is user-definable; in our implementation, we use: 
\begin{equation}
    \Delta(\bar{o}_a, \bar{o}_b) = 
        \max_{i \in \mathcal{O}_a \cap \mathcal{O}_b}
            \left[\lVert \mathbf{p}_a^{(i)} - \mathbf{p}_b^{(i)} \rVert_2\right]+
        \lambda \cdot \lvert \mathcal{O}_a \;\triangle\; \mathcal{O}_b \rvert
\end{equation}
where $\mathbf{p}^{(i)}$ is the 3D position of object $i$, $\mathcal{O}$ denotes the set of detected objects, $\triangle$ is the symmetric set difference, and $\lambda$ weights object appearance and disappearance relative to positional displacement. This check involves only Euclidean distances and set operations, adding negligible overhead.

\begin{figure}[t]
    \centering
    \includegraphics[width=0.95\columnwidth]{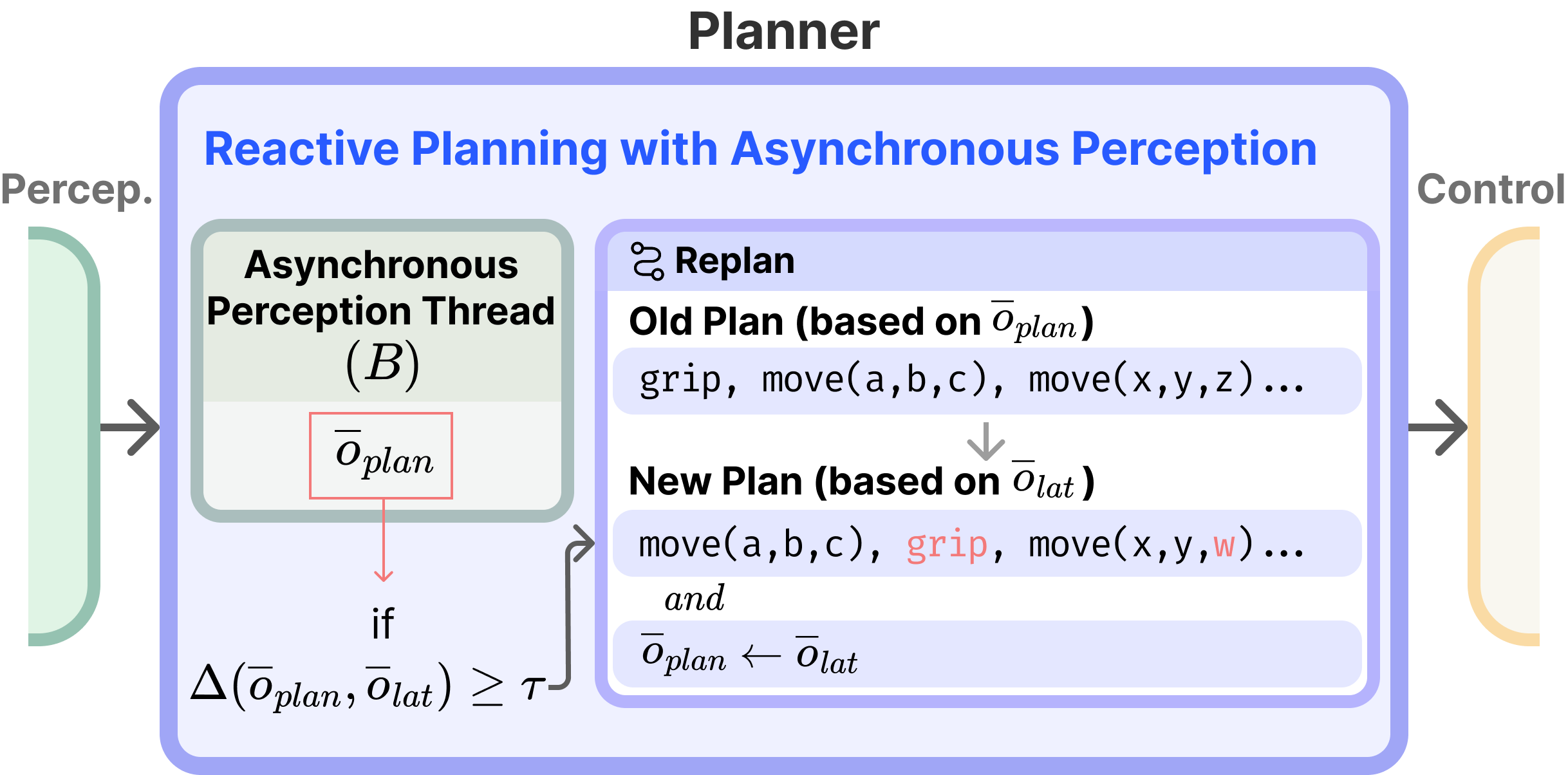}
    \caption{Reactive planning with asynchronous perception. The perception thread continuously updates a thread-safe buffer with the latest detection result. Replanning is triggered when the environment diverges materially from the current plan. This design preserves the high-level action sequence while regenerating subsequent primitive skills to reflect the current state.}
    \label{fig:Reactive Planning with Asynchronous Perception}
\end{figure}

As illustrated in Figure~\ref{fig:Reactive Planning with Asynchronous Perception}, when $\Delta(\bar{o}_{plan}, \bar{o}_{lat}) \ge \tau$, the high-level action sequence (e.g., \textsc{pick}~$\rightarrow$~\textsc{place} in pick-and-place tasks) is preserved while the primitive skills for the current and subsequent actions are regenerated using $\bar{o}_{lat}$. This avoids redundant high-level reasoning while updating spatial parameters to reflect the current state.

\subsection{Primitive skill fine-tuning}
\label{sec:gpr}

When a VLA serves as the controller, its single set of weights handles all manipulation phases indiscriminately, from approaching to grasping to releasing.
We decouple these phases by factoring manipulation into domain-invariant \emph{primitive skills}:
\begin{equation}
  \mathcal{P} = \{\textsc{move},\; \textsc{grip},\; \textsc{rotate}, ...\},
  \label{eq:primitive-set}
\end{equation}
Given a language instruction, the Planner decomposes the task into an ordered sequence $(p_1, \dots, p_T)$ with $p_t \in \mathcal{P}$, and each primitive is executed by a dedicated fine-tuned adapter rather than the monolithic backbone.


\textbf{Primitive-specific LoRA adapters.}
We attach a lightweight Low-Rank Adaptation (LoRA) module~\cite{hu2022lora} to a frozen VLA backbone $f_\theta$. Each adapter $\Delta\theta_k$ is trained on demonstrations filtered to that primitive:
\begin{equation}
  f_{\theta + \Delta\theta_k} : (i,\,\mathbf{s}_t,\,p_t) \mapsto \mathbf{a}_t^{(k)},
  \label{eq:adapter-inference}
\end{equation}
where $i$ is the detailed instruction, $\mathbf{s_t}$ is the concatenation of the current observation and robot state in time step $t$, and $p_t$ is the current primitive skill label included in plan $\bar{\mathcal{P}}$.

\newcommand{\RightComment}[1]{%
  \Statex \hfill$\triangleright$%
  \parbox[t]{0.46\linewidth}{\raggedright\footnotesize #1}%
}

\begin{algorithm}[t]
\vspace{0.05in}
\caption{\textsc{RoboBridge} Pipeline}
\label{alg:robobridge}

\begin{algorithmic}[1]

\Statex \textbf{Modules:}
\Statex \quad \textsc{Perceptor}: $D_\text{percept}$ (scene model), $\mathcal{B}$ (async buffer)
\Statex \quad \textsc{Planner}: $\textsc{Plan}(\cdot)$, $\textsc{Replan}(\cdot)$
\Statex \quad \textsc{Controller}: $f_\theta$ (VLA), $\{\Delta\theta_k\}$ (LoRA adapters)
\Statex \quad \textsc{Monitor}: $D_\text{failure}$ (lightweight success checker),
\Statex \quad \quad \quad \quad $D_\text{analyze}$ (failure diagnosis model)
\Statex \quad \textsc{Robot Interface}: $\textsc{env}$ (action interface)

\Procedure{RunTask}{$\textsc{env}$, instruction $i$}

    \State $o_0 \gets \textsc{env.Reset}()$
    \State $\bar{o}_{\text{plan}} \gets D_\text{percept}(o_0)$
    \State $\bar{\mathcal{P}} \gets \textsc{Plan}(i,\bar{o}_{\text{plan}})$
    \State launch $D_\text{percept}$ as async thread $\rightarrow \mathcal{B}$

    \For{each primitive $p_j \in \bar{\mathcal{P}}$}

        \State $\Delta\theta^* \gets \textsc{Resolve}(p_j)$ \Comment{Eq.~(6)}

        \While{$p_j$ not done}

            \State $a_t \gets f_{\theta+\Delta\theta^*}(i,\,\mathbf{s}_t,\,p_j)$
            \Comment{Eq.~(5)}
            \State $o_{t+1} \gets \textsc{env.Step}(a_t)$

            \State $(\textit{suc},\textit{conf}) \gets D_\text{failure}(o_t,p_j)$ \Comment{Eq.~(1)}

            \If{$\neg\,\textit{suc}$ \textbf{and} $\textit{conf} \ge \gamma$}
                \State $\textsc{env.Stop}()$
                \State $r \gets D_\text{analyze}(o_t,p_j)$ \Comment{Eq.~(2)}
                \State $\textsc{Recover}(r,\bar{\mathcal{P}},p_j)$
            \EndIf

        \EndWhile

        \State $\bar{o}_{\text{lat}} \gets \mathcal{B}.\textsc{Latest}()$

        \If{$\Delta(\bar{o}_{\text{plan}},\bar{o}_{\text{lat}}) \ge \tau$} \Comment{Eq.~(3)}
            \State $\bar{o}_{\text{plan}} \gets \bar{o}_{\text{lat}}$
            \State $\bar{\mathcal{P}} \gets \textsc{Replan}(i,\bar{o}_{\text{plan}})$
        \EndIf

    \EndFor

\EndProcedure
\end{algorithmic}
\vspace{-0.1in}
\end{algorithm}

\textbf{Controller switching.}
The controller maps each primitive to a dedicated policy and switches between them at execution time.
The framework is agnostic to the form of each policy; any action generator that conforms to the primitive interface can be registered.
For example, when each policy is a LoRA adapter on the frozen backbone $f_\theta$, switching reduces to a two-tier resolver: it selects the adapter associated with the requested primitive when available, and otherwise falls back to an adapter obtained by averaging all registered LoRA adapters.
\begin{equation}
  \textsc{Resolve}(p_t)
  =
  \begin{cases}
    \Delta\theta_{p_t} & \text{if adapter exists},\\
    \frac{1}{|\mathcal{P}|}\sum_{p \in \mathcal{P}} \Delta \theta_{p}
    & \text{otherwise}.
  \end{cases}
  \label{eq:adapter-resolve}
\end{equation}
Swapping LoRA modules is performed in-place without reloading the backbone, while classical IK controllers or scripted routines could equally serve as primitive policies.


\textbf{Action generation and execution.}
The selected adapter predicts a 7-DoF delta action:
\begin{equation}
  \mathbf{a}_t
  = \bigl[\,\delta\mathbf{x},\;\delta\boldsymbol{\phi},\;g\,\bigr]
  = f_{\theta+\Delta\theta^*}(i,\,\mathbf{s}_t,\,p_t),
  \label{eq:vla-output}
\end{equation}
where $\delta\mathbf{x} \in \mathbb{R}^3$ and $\delta\boldsymbol{\phi} \in \mathbb{R}^3$ are translational and rotational deltas, and $g \in [-1,1]$ is the gripper command. For \textsc{move}, the system applies $(\delta\mathbf{x}, \delta\boldsymbol{\phi})$ while holding the gripper fixed; for \textsc{grip}, the arm remains stationary and only $g$ is actuated.

The delta is dispatched as a Cartesian velocity command at control rate $f_c$:
\begin{equation}
  \mathbf{u}_t = \frac{1}{\Delta t}
  \begin{bmatrix}\delta\mathbf{x}\\ \delta\boldsymbol{\phi}\end{bmatrix},
  \quad \Delta t = 1/f_c.
  \label{eq:cartesian-cmd}
\end{equation}
When an inverse-kinematics (IK) solver is available, we convert the delta to an absolute target and solve for joint positions:
\begin{equation}
  \mathbf{x}_{t+1}^{\mathrm{ee}} = \mathbf{x}_t^{\mathrm{ee}} + \delta\mathbf{x},
  \qquad
  \mathbf{q}^* = \textsc{IK}\!\bigl(\mathbf{x}_{t+1}^{\mathrm{ee}},\;\mathbf{q}_t^{\mathrm{ee}}\bigr),
  \label{eq:ik-solve}
\end{equation}
falling back to Cartesian velocity commands if IK fails. This two-path execution strategy enables the same controller to operate across robots with different kinematic interfaces.

\section{EXPERIMENTS}
\subsection{Experiment Settings}
\textbf{Environments and Datasets.}
We evaluate \oursol in two settings: simulation and real-world manipulation. For simulation, we use LIBERO~\cite{liu2023libero} and RoboCasa~\cite{robocasa2024}.
LIBERO is a standardized benchmark suite for language-conditioned robotic manipulation that provides multiple task suites with diverse object interactions and goal specifications, enabling consistent evaluation of generalization across tasks and environments.
We follow the official evaluation protocol and use the provided human demonstration datasets for training and testing.
RoboCasa is a challenging large-scale kitchen simulation suite built for everyday manipulation. We follow the atomic-task benchmark with 24 tasks and train with 50 human demonstrations per task (1,200 total), collected via SpaceMouse teleoperation on a Franka Emika Research 3 arm mounted on an Omron mobile base.
For real-world experiments, we deploy on two robot platforms (Franka Emika Research 3, UR7e) for long-horizon, complex tasks.

\textbf{Evaluation Metric.}
We report \textbf{Success Rate (SR)}, defined as the fraction of rollouts that satisfy the task-completion predicate. In LIBERO and RoboCasa, success is determined by the simulator-provided condition for each atomic task. In real-world trials, an episode is successful if the robot achieves the task goal without human intervention.

\textbf{Baselines.}
We compare three VLA backbones:  
\textbf{SmolVLA}~\cite{shukor2025smolvla}, a lightweight VLA for compute-efficient deployment; \textbf{$\pi_{0.5}$}~\cite{black2025pi05}, a VLA co-trained across heterogeneous tasks for open-world generalization; and 
\textbf{GR00T-N1.5}~\cite{nvidia2025grootn15}, a state-of-the-art open foundation model for generalist robot control. For each backbone, we compare standalone deployment (\textbf{w/o}) against deployment within our framework (\textbf{w/RB}).

\textbf{Implementation Details.}
Each baseline VLA serves as the \emph{Controller}. For GR00T-N1.5-3B, we apply LoRA-only adaptation (rank 128, alpha 256, dropout 0.1) to the DiT action head and the vision projector, training separate adapters for \textsc{Move} and \textsc{Grip}. Both adapters use AdamW in bf16 with batch size 64, zero weight decay, and a cosine scheduler with 5\% warmup. Each adapter is trained for 500 epochs at a learning rate of $5\times10^{-5}$.
At inference time, rollouts use chunk stride 4, EMA smoothing ($\alpha{=}0.6$), 10 denoising steps, resolution $224{\times}224$, and a maximum horizon of 1000 steps.
The \emph{Perceptor} is fixed to Florence-2~\cite{xiao2024florence2} across all main-table experiments. The \emph{Planner} and the Phase-2 \emph{Monitor} are implemented with Claude Opus 4.6. Phase-1 monitoring uses a lightweight failure detector operating at approximately 5Hz outside the control loop, and thus does not block controller execution. The Planner operates at temperature 0.3 with reactive planning.
To ensure a fair comparison, we fix the training data, task set, rollout horizon, and all framework-side settings across models, varying only whether the controller runs standalone or within our framework.

\begin{table}[t]
\centering
\vspace{0.05in}
\caption{Performance comparison in LIBERO.}
\vspace{-0.05in}
\label{tab:libero}
\small
\setlength{\tabcolsep}{2.6pt}
\renewcommand{\arraystretch}{1.05}
\resizebox{0.99\columnwidth}{!}{%
\begin{tabular}{@{}l|cc|cc|cc|cc@{}}
\toprule
\textbf{Task}
& \multicolumn{2}{c}{LIBERO-Object}
& \multicolumn{2}{c}{LIBERO-Spatial}
& \multicolumn{2}{c}{LIBERO-Goal}
& \multicolumn{2}{c}{LIBERO-Long}\\
\cmidrule(lr){2-3}\cmidrule(lr){4-5}\cmidrule(lr){6-7}\cmidrule(lr){8-9}
& w/o & w/RB
& w/o & w/RB
& w/o & w/RB
& w/o & w/RB \\
\midrule

Task01& 23.5\%  & 47.1\% & 94.1\%  & 94.1\% & 100\%  & 100\% & 5.9\%  & 35.3\% \\
Task02& 0.0\%  & 0.0\% & 94.1\%  & 76.5\% & 100\%  & 94.1\% & 17.6\%  & 41.2\% \\
Task03& 0.0\%  & 0.0\% & 88.2\%  & 82.4\% & 70.6\%  & 76.5\% & 41.2\%  & 41.2\% \\
Task04& 0.0\%  & 0.0\% & 82.4\%  & 82.4\% & 58.8\%  & 64.7\% & 0.0\%  & 0.0\% \\
Task05& 5.9\%  & 11.8\% & 82.4\%  & 82.4\% & 41.2\%  & 58.8\% & 0.0\%  & 17.6\% \\
Task06& 0.0\%  & 5.9\% & 76.5\%  & 70.6\% & 70.6\%  & 47.1\% & 5.9\%  & 11.8\% \\
Task07& 5.9\%  & 11.8\% & 70.6\%  & 64.7\% & 41.2\%  & 47.1\% & 0.0\%  & 5.9\% \\
Task08& 0.0\%  & 0.0\% & 47.1\%  & 52.9\% & 29.4\%  & 35.3\% & 35.3\%  & 41.2\% \\
Task09& 5.9\%  & 11.8\% & 41.2\%  & 47.1\% & 5.9\%  & 17.6\% & 0.0\%  & 5.9\% \\
Task10& 5.9\%  & 11.8\% & 47.1\%  & 82.4\% & 23.5\%  & 11.8\% & 0.0\%  & 0.0\% \\

\midrule
\textbf{Avg.} & \textbf{4.7\%} & \textbf{10.0\%} & \textbf{72.4\%} & \textbf{73.5\%} & \textbf{54.1\%} & \textbf{55.3\%} & \textbf{10.6\%} & \textbf{20.0\%} \\
\textbf{$\Delta$ Avg.}
& \multicolumn{2}{c|}{\textbf{+5.3\%}}
& \multicolumn{2}{c|}{\textbf{+1.1\%}}
& \multicolumn{2}{c|}{\textbf{+1.2\%}}
& \multicolumn{2}{c}{\textbf{+9.4\%}} \\
\bottomrule
\end{tabular}%
}
\vspace{-0.1in}

\end{table}

\begin{table}[t]
\centering
\vspace{0.05in}
\caption{Performance comparison in RoboCasa.}
\vspace{-0.05in}
\label{tab:robocasa}
\small
\setlength{\tabcolsep}{2.6pt}
\renewcommand{\arraystretch}{1.05}
\resizebox{0.99\columnwidth}{!}{%
\begin{tabular}{@{}>{\scriptsize\raggedright\arraybackslash}l|cc|cc|cc@{}}
\toprule
\textbf{Task}
& \multicolumn{2}{c}{SmolVLA}
& \multicolumn{2}{c}{\scalebox{1.3}{$\pi_{0.5}$}}
& \multicolumn{2}{c}{GR00T-N1.5} \\
\cmidrule(lr){2-3}\cmidrule(lr){4-5}\cmidrule(lr){6-7}
& w/o & w/RB
& w/o & w/RB
& w/o & w/RB \\
\midrule

CloseDoubleDoor & 0.0\%  & 0.0\% & 0.0\%  & 0.0\% & 0.0\%  & 0.0\% \\
CloseSingleDoor & 0.0\%  & 47.1\% & 0.0\%  & 0.0\% & 5.9\%  & 5.9\% \\
OpenDoubleDoor & 0.0\%  & 0.0\% & 0.0\%  & 5.9\% & 0.0\%  & 0.0\% \\
OpenSingleDoor & 0.0\%  & 29.4\% & 0.0\%  & 0.0\% & 0.0\%  & 0.0\% \\
CloseDrawer & 23.5\%  & 23.5\% & 0.0\%  & 0.0\% & 17.6\% & 70.6\% \\
OpenDrawer & 0.0\%  & 0.0\% & 0.0\%  & 0.0\% & 0.0\%  & 0.0\% \\
CoffeePressButton & 0.0\%  & 0.0\% & 0.0\%  & 23.5\% & 17.6\%  & 23.5\% \\
CoffeeServeMug & 0.0\%  & 0.0\% & 0.0\% & 0.0\% & 0.0\%  & 5.9\% \\
CoffeeSetupMug & 0.0\%  & 0.0\% & 0.0\%  & 0.0\% & 0.0\%  & 0.0\% \\
PnPCabToCounter & 0.0\%  & 0.0\% & 29.4\%  & 0.0\% & 0.0\%  & 0.0\% \\
PnPCounterToCab & 0.0\%  & 0.0\% & 0.0\%  & 0.0\% & 0.0\%  & 0.0\% \\
PnPCounterToMicrowave & 0.0\%  & 0.0\% & 0.0\%  & 0.0\% & 0.0\%  & 0.0\% \\
PnPCounterToSink & 0.0\%  & 0.0\% & 0.0\%  & 0.0\% & 0.0\%  & 0.0\% \\
PnPCounterToStove & 0.0\%  & 0.0\% & 0.0\%  & 0.0\% & 0.0\%  & 0.0\% \\
PnPMicrowaveToCounter & 0.0\%  & 0.0\% & 0.0\%  & 0.0\% & 0.0\%  & 0.0\% \\
PnPSinkToCounter & 0.0\%  & 0.0\% & 0.0\%  & 0.0\% & 0.0\%  & 0.0\% \\
PnPStoveToCounter & 0.0\%  & 0.0\% & 0.0\%  & 0.0\% & 0.0\%  & 0.0\% \\
TurnOffMicrowave & 17.6\%  & 0.0\% & 0.0\%  & 52.9\% & 0.0\%  & 11.8\% \\
TurnOnMicrowave & 5.9\% & 0.0\% & 23.5\% & 0.0\% & 0.0\%  & 0.0\% \\
TurnOffSinkFaucet & 29.4\%  & 17.6\% & 5.9\%  & 23.5\% & 23.5\%  & 52.9\% \\
TurnOnSinkFaucet & 0.0\%  & 0.0\% & 0.0\%  & 5.9\% & 0.0\%  & 0.0\% \\
TurnSinkSpout & 5.9\% & 0.0\% & 52.9\%  & 29.4\% & 35.3\% & 47.1\% \\
TurnOffStove & 0.0\%  & 5.9\% & 0.0\%  & 0.0\% & 0.0\%  & 5.9\% \\
TurnOnStove & 0.0\%  & 41.2\% & 0.0\%  & 0.0\% & 0.0\%  & 11.8\% \\

\midrule
\textbf{Avg.} & \textbf{3.4\%} & \textbf{6.9\%} & \textbf{3.4\%} & \textbf{5.9\%}& \textbf{4.2\%} & \textbf{9.8\%} \\
\textbf{$\Delta$ Avg.}
& \multicolumn{2}{c|}{\textbf{+3.5\%}}
& \multicolumn{2}{c|}{\textbf{+2.5\%}}
& \multicolumn{2}{c}{\textbf{+5.6\%}} \\
\midrule
\textbf{Avg. w/o PnP} & \textbf{5.1\%} & \textbf{10.7\%} & \textbf{7.4\%} & \textbf{8.8\%}& \textbf{6.2\%} & \textbf{14.7\%} \\
\textbf{$\Delta$ Avg. w/o PnP}
& \multicolumn{2}{c|}{\textbf{+5.5\%}}
& \multicolumn{2}{c|}{\textbf{+1.4\%}}
& \multicolumn{2}{c}{\textbf{+8.5\%}} \\
\bottomrule
\end{tabular}%
}
\end{table}

\subsection{Main Results}

\textbf{Simulation results.}
We evaluate \oursol on two simulation benchmarks, LIBERO and RoboCasa, using three VLA backbones: GR00T-N1.5 for LIBERO, and GR00T-N1.5 together with SmolVLA and $\pi_{0.5}$ for RoboCasa. Tables~\ref{tab:libero} and~\ref{tab:robocasa} report results under a controlled comparison where we keep the controller fixed and vary only whether it is deployed standalone or wrapped by \oursol. On LIBERO, standalone VLAs already achieve moderate success, yet \oursol still yields a consistent gain, improving average task success from 35.5\% to 39.7\%. RoboCasa is substantially more challenging due to longer horizons and frequent execution drift, where standalone performance collapses and many tasks fail consistently; wrapping the same controllers with \oursol improves performance across all backbones and a broad subset of tasks, raising the average success from 3.7\% to 7.5\% (and from 6.2\% to 11.4\% when excluding pick-and-place tasks).
These improvements arise without scaling or retraining the base policies in the standalone versus framework comparison. Instead, \oursol supplies agent capabilities that monolithic action predictors lack: two-phase monitoring detects failures early and applies hierarchical recovery to prevent cascading errors, while reactive planning with asynchronous perception reduces brittleness to scene drift by updating spatial parameters only when divergence is detected. When primitive-specific adapters are available, controller switching further reduces interference across manipulation phases. Together, these mechanisms explain why \oursol improves performance not only on long-horizon tasks in RoboCasa, but also on LIBERO where standalone controllers are already reasonably capable.

\begin{figure*}[t]
    \centering
    \includegraphics[width=0.95\textwidth]{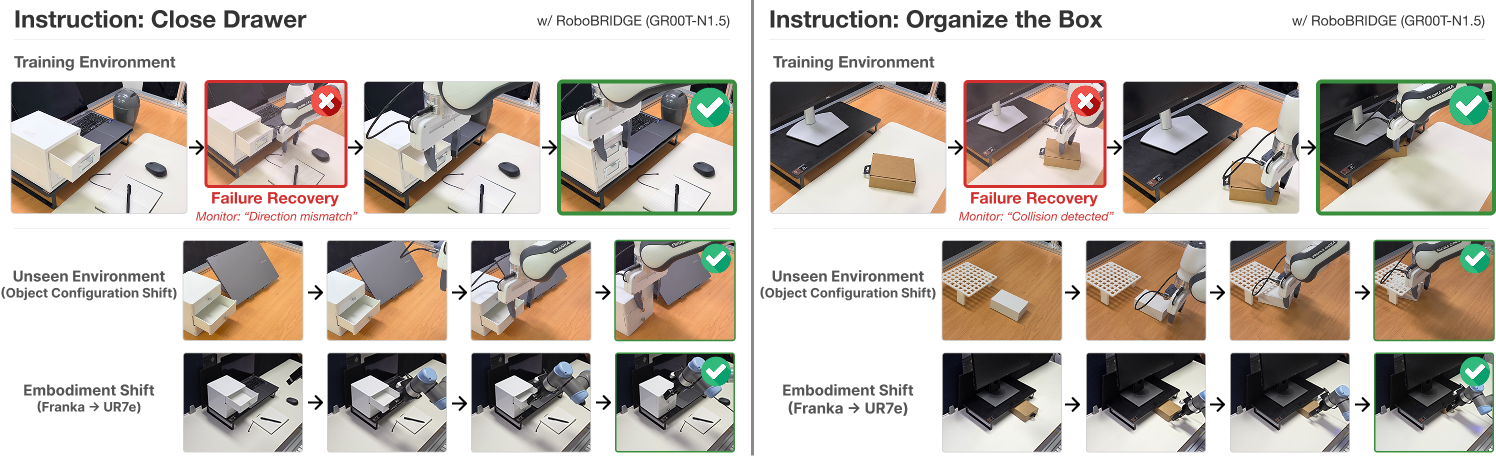}
    \caption{Real-world execution traces with \oursol (GR00T-N1.5 controller). Each task is shown in the training environment (top) and an unseen environment (bottom). The Monitor detects execution failures and triggers recovery during rollout (red boxes), enabling the agent to resume progress and complete the task (green boxes). The same monitoring and recovery behavior transfers across environment changes without additional adaptation.}
    \label{fig:fig5}
    \vspace{-0.1in}
\end{figure*}

\textbf{Real-world case study.}
We deploy \oursol on a Franka Emika Research 3 arm to validate its applicability beyond simulation.
Figure~\ref{fig:fig5} shows representative execution traces in both training environments (top row) and unseen environments (bottom row).
Red boxes mark the moment when the Monitor flags a failure and triggers recovery, while green boxes indicate successful completion.
Across these episodes, \oursol detects execution misalignment early, intervenes with a minimal recovery action, and resumes progress without restarting the full plan. This prevents local manipulation errors from cascading into downstream steps, which is especially important in longer sequences where a single mistake can derail the remainder of the rollout. The same monitoring and recovery behavior transfers to unseen environments without additional adaptation, demonstrating robustness to visual variation and actuation uncertainty.

\begin{table}[t]
\centering
\caption{Effect of LLM backbone and two-phase monitoring}
\label{tab:planner_ablation}
\footnotesize
\begin{tabular}{l|ccc}
\toprule
\textbf{Planner} & \textbf{w/o Monitor} & \textbf{w/ Monitor} & \textbf{$\Delta$} \\
\midrule
Claude Opus 4.6   & 6.6\%  & \textbf{14.7\%} & +8.1\% \\
Claude Sonnet 4.6  & 8.0\%  & 9.8\%  & +1.8\% \\
Claude Haiku 4.5 & 6.3\%  & 6.3\%  & +0.0\% \\
\midrule
GPT-5 mini        & 2.7\%  & 8.9\%  & +6.2\% \\
GPT-5 nano        & 2.7\%  & 5.4\%  & +2.7\% \\
\midrule
Gemini-3.1 Pro    & 6.0\%  & 8.0\%  & +2.0\% \\
Gemini-3 Flash    & 1.8\%  & 2.7\%  & +0.9\% \\
\bottomrule
\end{tabular}
\end{table}

\subsection{Analysis}
\textbf{Effect of LLM backbone and monitoring.}
Table~\ref{tab:planner_ablation} examines how the choice of LLM backbone and the presence of two-phase monitoring jointly affect framework performance, using GR00T-N1.5 on RoboCasa (PnP tasks excluded).
Without monitoring, all backbones fall within a narrow 1.8--8.0\% range, indicating that planning quality alone is insufficient to bridge the gap between a VLA and a reliable agent.
Enabling monitoring reveals a clear separation: Claude Opus 4.6 achieves the highest gain (+8.1\%, reaching 14.7\%), followed by GPT-5 mini (+6.2\%).
In contrast, smaller models such as Gemini-3 Flash and Claude Haiku 4.5 show only marginal improvements, suggesting that effective failure diagnosis requires sufficient reasoning capacity in the backbone.
These results demonstrate that neither planning nor monitoring alone is sufficient; both are necessary, and their combined benefit scales with the backbone's diagnostic ability.

\begin{table}[th]
\centering
\caption{Analysis for controller}
\label{tab:2}

\small 
\setlength{\tabcolsep}{4pt}
\renewcommand{\arraystretch}{1.08} 

\resizebox{0.99\columnwidth}{!}{%
\begin{tabular}{>{\footnotesize\raggedright\arraybackslash}l|cc|cc|c|c}
\toprule
\textbf{Task}
& \multicolumn{2}{c|}{LoRA FT}
& \multicolumn{2}{c|}{Full FT}
& {IK}
& {CycleVLA}\\
\cmidrule(lr){2-3}\cmidrule(lr){4-5}\cmidrule(lr){6-6}\cmidrule(lr){7-7}
& w/o & w/RB
& w/o & w/RB
& w/RB & w/o\\
\midrule
CoffeePressButton & 17.6\% & 23.5\% & 41.2\% & 41.2\% & 5.9\% & 0.0\%\\
TurnOffMicrowave & 0.0\% & 11.8\% & 23.5\% & 35.3\% & 29.4\% & 0.0\%\\
TurnOffSinkFaucet & 23.5\% & 52.9\% & 29.4\% & 41.2\% & 11.8\% & 23.5\%\\
TurnSinkSpout & 35.3\% & 47.1\% & 59.7\% & 64.7\% & 41.2\% & 14.3\%\\
PnPCabToCounter & 0.0\% & 0.0\% & 5.9\% & 17.6\% & 0.0\% & 0.0\%\\
\midrule
\textbf{Avg.}
& \textbf{15.3\%} & \textbf{27.1\%}
& \textbf{31.9\%} & \textbf{40.0\%}
& \textbf{22.1\%} & \textbf{7.5\%}\\
\textbf{$\Delta$ Avg.}
& \multicolumn{2}{c|}{\textbf{+11.8\%}}
& \multicolumn{2}{c|}{\textbf{+8.1\%}}
& \textbf{---}
& \textbf{---}\\
\bottomrule
\end{tabular}%
}
\end{table}

\textbf{Analysis of controller.}
Table~\ref{tab:2} compares four controller configurations on a subset of five RoboCasa tasks: LoRA fine-tuning (LoRA FT), full fine-tuning (Full FT), a classical inverse-kinematics (IK) controller, and CycleVLA~\cite{cyclevla2026}, a recent VLA variant that integrates subtask backtracking and self-correction decoding into the model itself.
Since the IK controller relies on the Planner to specify target poses, it operates only within \oursol and is not evaluated standalone.
\oursol improves both fine-tuning strategies, with average success rates rising by +11.8 and +8.1 percentage points respectively, confirming that the framework generalizes across controller types.
Most notably, LoRA fine-tuning combined with \oursol achieves 27.1\%, approaching the standalone full fine-tuning baseline (31.9\%) despite updating far fewer parameters.
Full fine-tuning itself further benefits from \oursol, reaching 40.0\%, which shows that even strong controllers gain from structured monitoring and replanning.
The IK controller achieves 22.1\% within \oursol, demonstrating that the framework's policy-agnostic design extends beyond learned policies to classical controllers.
CycleVLA, which augments a VLA with subtask backtracking and self-correction decoding, averages only 7.5\% without our framework, suggesting that model-internal correction alone is insufficient compared to the external orchestration \oursol provides.

\textbf{Failure cases analysis.}
We identify two dominant failure modes across our experiments.
First, \textit{perception errors} occur when the Perceptor misidentifies or mislocalizes target objects, causing downstream primitives to execute with incorrect object poses. These errors are most common in cluttered scenes with occlusions or visually similar objects.
\begin{figure}[t]
    \centering
    \includegraphics[width=\columnwidth]{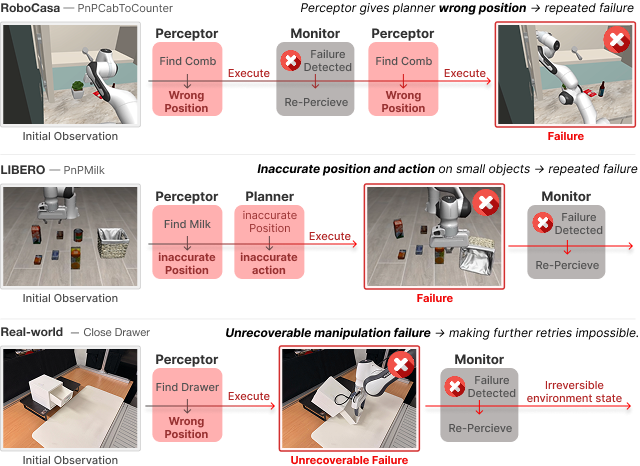}
    \caption{Observation of two dominant failure modes (perception errors and unrecoverable manipulation failures) in simulation and real-world. These cases highlight the need to improve reliability by strengthening perception robustness under occlusion and introduction of explicit verification and validation for contact rich behaviors.}
    \label{fig:fig6}
    \vspace{-0.1in}
\end{figure}
Second, \textit{unrecoverable manipulation failures} arise in contact-rich
interactions where execution errors irreversibly alter the environment
state. In such cases, further retries become futile because the object
configuration no longer permits successful completion.
These cases highlight the need for stronger verification and validation mechanisms that can (i) detect when the current controller is unlikely to succeed, (ii) select alternative strategies or controllers when available, and (iii) provide calibrated stopping criteria to avoid wasting steps on futile retries.
Overall, these failure modes suggest that improving perception robustness under occlusion and introducing explicit verification and validation for contact rich behaviors are key directions for improving reliability.

\section{CONCLUSION}
We presented \oursol, a modular, generalizable orchestration framework that converts any action-generating policy into a reliable robotic agent. By coordinating five modules, namely Monitor, Perceptor, Planner, Controller, and Robot Interface, with two-phase hierarchical recovery and reactive asynchronous planning, \oursol enables robust operation in dynamic real-world settings without retraining the base policy. When a VLA serves as the controller, primitive skill fine-tuning with dedicated LoRA adapters further reduces sensitivity to domain shifts. Evaluated on 24 RoboCasa tasks and real-world trials across two robot platforms with three VLA backbones, \oursol yields consistent improvements over standalone baselines, demonstrating that a general orchestration architecture can systematically bridge the gap between learned policies and robust deployment.

\textbf{Limitations and future work.}
Our primitive skill vocabulary covers many single arm tabletop behaviors, but contact rich interactions, deformable objects, and bimanual tasks will require richer primitives and state representations. Monitoring thresholds and recovery rules are currently set manually; learning these from interaction data is a promising direction. A key next step is to add explicit verification and validation layers that can assess feasibility, detect low likelihood of success, and trigger alternative strategies or safe termination.  Broadly, we envision \oursol as a framework that unifies data collection, verification, validation, and deployment under a single orchestration architecture.








\bibliographystyle{IEEEtran}
\bibliography{ref}

@inproceedings{cai2024latm,
  title     = {Large Language Models as Tool Makers},
  author    = {Cai, Tianle and Wang, Xuezhi and Ma, Tengyu and Chen, Xinyun and Zhou, Denny},
  booktitle = {International Conference on Learning Representations},
  year      = {2024}
}

@inproceedings{wu2024avatar,
  title     = {AvaTaR: Optimizing LLM Agents for Tool Usage via Contrastive Reasoning},
  author    = {Wu, Shirley and Zhao, Shiyu and Huang, Qian and Huang, Kexin and Yasunaga, Michihiro and Cao, Kaidi and Ioannidis, Vassilis N. and Subbian, Karthik and Leskovec, Jure and Zou, James},
  booktitle = {Advances in Neural Information Processing Systems},
  year      = {2024}
}

@inproceedings{kim2024openvla,
  title     = {{OpenVLA}: An Open-Source Vision-Language-Action Model},
  author    = {Kim, Moo Jin and Pertsch, Karl and Karamcheti, Siddharth and Xiao, Ted and Balakrishna, Ashwin and Nair, Suraj and Rafailov, Rafael and Foster, Ethan and others},
  booktitle = {Proceedings of The 8th Conference on Robot Learning},
  pages     = {2679--2713},
  year      = {2025}
}

@article{shukor2025smolvla,
  title     = {{SmolVLA}: A Vision-Language-Action Model for Affordable and Efficient Robotics},
  author    = {Shukor, Mustafa and Aubakirova, Dana and Capuano, Francesco and Kooijmans, Pepijn and Palma, Steven and Zouitine, Adil and Aractingi, Michel and Pascal, Caroline and others},
  journal = {arXiv preprint arXiv:2506.01844},
  year      = {2025},
}

@inproceedings{black2025pi05,
  title     = {$\pi_{0.5}$: a Vision-Language-Action Model with Open-World Generalization},
  author    = {Black, Kevin and Brown, Noah and Darpinian, James and Dhabalia, Karan and Driess, Danny and Esmail, Adnan and Equi, Michael Robert and Finn, Chelsea and others },
  booktitle = {Proceedings of The 9th Conference on Robot Learning},
  pages     = {17--40},
  year      = {2025},
}

@misc{nvidia2025grootn15,
  title        = {{NVIDIA Isaac GR00T N1.5} Model Card ({GR00T-N1.5-3B})},
  author       = {{NVIDIA}},
  year         = {2025},
  url          = {https://huggingface.co/nvidia/GR00T-N1.5-3B},
  note         = {Accessed: 2026-02-09}
}

@inproceedings{robocasa2024,
  title     = {RoboCasa: Large-Scale Simulation of Everyday Tasks for Generalist Robots},
  author    = {Nasiriany, Soroush and Maddukuri, Abhiram and Zhang, Lance and Parikh, Adeet and Lo, Aaron and Joshi, Abhishek and Mandlekar, Ajay and Zhu, Yuke},
  booktitle = {Proceedings of Robotics: Science and Systems},
  year      = {2024},
}

@inproceedings{hu2022lora,
  title     = {{LoRA}: Low-Rank Adaptation of Large Language Models},
  author    = {Hu, Edward J. and Shen, Yelong and Wallis, Phillip and Allen-Zhu, Zeyuan and Li, Yuanzhi and Wang, Shean and Wang, Lu and Chen, Weizhu},
  booktitle = {International Conference on Learning Representations},
  year      = {2022}
}

@inproceedings{openx2024,
  title     = {Open X-Embodiment: Robotic Learning Datasets and RT-X Models},
  author    = {{Open X-Embodiment Collaboration}},
  booktitle = {2024 IEEE International Conference on Robotics and Automation},
  pages     = {6892--6903},
  year      = {2024},
  doi       = {10.1109/ICRA57147.2024.10611477},
}

@inproceedings{octo2024,
  title     = {Octo: An Open-Source Generalist Robot Policy},
  author    = {Ghosh, Dibya and Walke, Homer Rich and Pertsch, Karl and Black, Kevin and others},
  booktitle = {Proceedings of Robotics: Science and Systems},
  year      = {2024},
}

@inproceedings{droid2024,
  title     = {DROID: A Large-Scale In-the-Wild Robot Manipulation Dataset},
  author    = {Khazatsky, Alexander and Pertsch, Karl and Nair, Suraj and others},
  booktitle = {Proceedings of Robotics: Science and Systems},
  year      = {2024},
}

@inproceedings{zhou2024llmbt,
  title     = {LLM-BT: Performing Robotic Adaptive Tasks based on Large Language Models and Behavior Trees},
  author    = {Zhou, Haotian and Lin, Yunhan and Yan, Longwu and Zhu, Jihong and Min, Huasong},
  booktitle = {2024 IEEE International Conference on Robotics and Automation},
  pages     = {16655--16661},
  year      = {2024},
  doi       = {10.1109/ICRA57147.2024.10610183},
}

@inproceedings{zhou2024isrllm,
  title     = {ISR-LLM: Iterative Self-Refined Large Language Model for Long-Horizon Sequential Task Planning},
  author    = {Zhou, Zhehua and Song, Jiayang and Yao, Kunpeng and Shu, Zhan and Ma, Lei},
  booktitle = {2024 IEEE International Conference on Robotics and Automation},
  pages     = {2081--2088},
  year      = {2024},
  doi       = {10.1109/ICRA57147.2024.10610065},
}

@article{brohan2022rt,
  title={Rt-1: Robotics transformer for real-world control at scale},
  author={Brohan, Anthony and Brown, Noah and Carbajal, Justice and Chebotar, Yevgen and Dabis, Joseph and Finn, Chelsea and Gopalakrishnan, Keerthana and Hausman, Karol and Herzog, Alex and Hsu, Jasmine and others},
  journal={arXiv preprint arXiv:2212.06817},
  year={2022}
}

@inproceedings{ahn2022can,
  title={Do as i can, not as i say: Grounding language in robotic affordances},
  author={Ahn, Michael and Brohan, Anthony and Brown, Noah and Chebotar, Yevgen and Cortes, Omar and David, Byron and Finn, Chelsea and Fu, Chuyuan and Gopalakrishnan, Keerthana and Hausman, Karol and others},
  booktitle={Proceedings of The 6th Conference on Robot Learning},
  pages={287--318},
  year={2023}
}

@inproceedings{zhou2025code,
  title={Code-as-monitor: Constraint-aware visual programming for reactive and proactive robotic failure detection},
  author={Zhou, Enshen and Su, Qi and Chi, Cheng and Zhang, Zhizheng and Wang, Zhongyuan and Huang, Tiejun and Sheng, Lu and Wang, He},
  booktitle={Proceedings of the IEEE/CVF Conference on Computer Vision and Pattern Recognition},
  pages={6919--6929},
  year={2025}
}

@inproceedings{guo2024doremi,
  title={Doremi: Grounding language model by detecting and recovering from plan-execution misalignment},
  author={Guo, Yanjiang and Wang, Yen-Jen and Zha, Lihan and Chen, Jianyu},
  booktitle={2024 IEEE/RSJ International Conference on Intelligent Robots and Systems},
  pages={12124--12131},
  year={2024},
  doi={10.1109/IROS58592.2024.10802284}
}

@inproceedings{huang2023voxposer,
  title={Voxposer: Composable 3d value maps for robotic manipulation with language models},
  author={Huang, Wenlong and Wang, Chen and Zhang, Ruohan and Li, Yunzhu and Wu, Jiajun and Fei-Fei, Li},
  booktitle={Proceedings of The 7th Conference on Robot Learning},
  pages={540--562},
  year={2023}
}

@article{santos2026alrm,
  title={ALRM: Agentic LLM for Robotic Manipulation},
  author={Santos, Vitor Gaboardi dos and Khadraoui, Ibrahim and Farhat, Ibrahim and Yous, Hamza and Teffahi, Samy and Hacid, Hakim},
  journal={arXiv preprint arXiv:2601.19510},
  year={2026}
}

@inproceedings{wu2024autogen,
  title={Autogen: Enabling next-gen LLM applications via multi-agent conversations},
  author={Wu, Qingyun and Bansal, Gagan and Zhang, Jieyu and Wu, Yiran and Li, Beibin and Zhu, Erkang and Jiang, Li and Zhang, Xiaoyun and Zhang, Shaokun and Liu, Jiale and others},
  booktitle={First Conference on Language Modeling},
  year={2024}
}

@inproceedings{safe2025,
    title = {SAFE: Multitask Failure Detection for Vision-Language-Action Models},
    author = {Gu, Qiao and Ju, Yuanliang and Sun, Shengxiang and Gilitschenski, Igor and Nishimura, Haruki and Itkina, Masha and Shkurti, Florian},
    booktitle = {Advances in Neural Information Processing Systems},
    year = {2025},
}

@article{failsafe2025,
title = {FailSafe: Reasoning and Recovery from Failures in Vision-Language-Action Models},
author = {Lin, Zijun and Duan, Jiafei and Fang, Haoquan and Fox, Dieter and Krishna, Ranjay and Tan, Cheston and Wen, Bihan},
journal = {arXiv preprint arXiv:2510.01642},
year = {2025},
}

@article{cyclevla2026,
    title = {CycleVLA: Proactive Self-Correcting Vision-Language-Action Models via Subtask Backtracking and Minimum Bayes Risk Decoding},
    author = {Ma, Chenyang and Yang, Guangyu and Lu, Kai and Xu, Shitong and Byrne, Bill and Trigoni, Niki and Markham, Andrew},
    journal = {arXiv preprint arXiv:2601.02295},
    year = {2026},
}

@article{kachaev2025don,
  title={Don't Blind Your VLA: Aligning Visual Representations for OOD Generalization},
  author={Kachaev, Nikita and Kolosov, Mikhail and Zelezetsky, Daniil and Kovalev, Alexey K and Panov, Aleksandr I},
  journal={arXiv preprint arXiv:2510.25616},
  year={2025}
}

@inproceedings{xiao2024florence2,
  title     = {Florence-2: Advancing a Unified Representation for a Variety of Vision Tasks},
  author    = {Xiao, Bin and Wu, Haiping and Xu, Weijian and Dai, Xiyang and Hu, Houdong and Lu, Yumao and Zeng, Michael and Liu, Ce and Yuan, Lu},
  booktitle = {Proceedings of the IEEE/CVF Conference on Computer Vision and Pattern Recognition},
  pages={4818-4829},
  year      = {2024}
}

@inproceedings{liu2023libero,
  title={LIBERO: Benchmarking Knowledge Transfer for Lifelong Robot Learning},
  author={Liu, Bo and Zhu, Yifeng and Gao, Chongkai and Feng, Yihao and Liu, Qiang and Zhu, Yuke and Stone, Peter},
  booktitle={Advances in Neural Information Processing Systems},
  pages={10700-10714},
  year={2023}
}

@article{zhong2025survey,
  title={A survey on vision-language-action models: An action tokenization perspective},
  author={Zhong, Yifan and Bai, Fengshuo and Cai, Shaofei and Huang, Xuchuan and Chen, Zhang and Zhang, Xiaowei and Wang, Yuanfei and Guo, Shaoyang and Guan, Tianrui and Lui, Ka Nam and others},
  journal={arXiv preprint arXiv:2507.01925},
  year={2025}
}

@article{sapkota2025vision,
  title={Vision-Language-Action (VLA) Models: Concepts, Progress, Applications and Challenges},
  author={Sapkota, Ranjan and Cao, Yang and Roumeliotis, Konstantinos I and Karkee, Manoj},
  journal={arXiv preprint arXiv:2505.04769},
  year={2025}
}

@misc{langchain_website,
  title  = {LangChain},
  author = {{LangChain}},
  url    = {https://www.langchain.com},
  note   = {Accessed: 2026-02-09}
}

\end{document}